\documentclass{ifacconf}

\usepackage{graphicx}      
\usepackage{natbib}        
\usepackage{amsmath}
\usepackage{siunitx}

\newtheorem{def1}{Definition}
\newtheorem{not1}{Notation}

\usepackage{url}

\begin{document}
\begin{frontmatter}

\title{Graceful User Following for Mobile Balance Assistive Robot in Daily Activities Assistance} 

\thanks[footnoteinfo]{This work is supported by SG Health Assistive and Robotics Programme  (SHARP) Grant - Development and POC of Care Assistant and  Rehabilitation Enabling (CARE) Robots (SERC 1922200003).}

\author[First]{Yifan Wang} 
\author[Second]{Meng Yuan} 
\author[Second]{Lei Li}
\author[Third]{Karen Sui Geok Chua}
\author[Third]{Seng Kwee Wee}
\author[Second]{Wei Tech Ang}

\address[First]{School of Mechanical and Aerospace Engineering, Nanyang Technological University, 639798 Singapore.} 
\address[Second]{Rehabilitation Research Institute of Singapore, Nanyang Technological University, 308232 Singapore \\(Corresponding e-mail: meng.yuan@ntu.edu.sg)}
\address[Third]{Tan Tock Seng Hospital, 308233 Singapore }

\begin{abstract}                
Numerous diseases and aging can cause degeneration of people's balance ability resulting in limited mobility and even high risks of fall. Robotic technologies can provide more intensive rehabilitation exercises or be used as assistive devices to compensate for balance ability. However, With the new healthcare paradigm shifting from hospital care to home care, there is a gap in robotic systems that can provide care at home. This paper introduces Mobile Robotic Balance Assistant (MRBA), a compact and cost-effective balance assistive robot that can provide both rehabilitation training and activities of daily living (ADLs) assistance at home. A three degrees of freedom (3-DoF) robotic arm was designed to mimic the therapist arm function to provide balance assistance to the user. To minimize the interference to users' natural pelvis movements and gait patterns, the robot must have a Human-Robot Interface(HRI) that can detect user intention accurately and follow the user's movement smoothly and timely. Thus, a graceful user following control rule was proposed. The overall control architecture consists of two parts: an observer for human inputs estimation and an LQR-based controller with disturbance rejection. The proposed controller is validated in high-fidelity simulation with actual human trajectories, and the results successfully show the effectiveness of the method in different walking modes.
\end{abstract}

\begin{keyword}
Human-machine systems, assistive robots, human following, disturbance observer, linear quadratic regulator
\end{keyword}

\end{frontmatter}

\section{Introduction}

Elderly people often face the degradation of their balance control ability which can be caused by natural aging of bodies or geriatric diseases like insults of the central nervous system such as spinal cord injury, cerebrovascular accident, Parkinson’s disease, peripheral neuropathies, etc. and chronic injuries of the muscle-skeleton system such as chronic ankle sprains, scoliosis, amputation, etc. according to \cite{rosso2013aging}. The weakened balance control ability will lead to an increased risk of fall, which is the main cause of accidental death in the elderly \cite{rubenstein2006falls}. 

Many balance assistive robots were designed for people with balance impairments to aid gait training and provide balance support on time. KineAssist introduced by \cite{peshkin2005kineassist}, was one of the early attempts that provide walking or running support according to specific body weight and functional training tasks. Similar overground walking trainers were developed, including Robotic Walker for Gait Rehabilitation introduced by \cite{mun2015development}, NaTUregait developed by \cite{wang2011synchronized} and GaitEnable delivered by \cite{morbi2012gaitenable}. However, these current products are more suitable to be applied in institutions or medical centers due to their large size and high cost. There is a gap in products for post-rehabilitation patients released from hospitals or rehabilitation centers to use in home and community scenarios to assist their activities of daily living (ADLs). Moreover, users’ natural gaits will be influenced by the inertial transmitted through the human-robot interface (HRI) due to the transparency issues which are critical in HRI design. 

User following is one of the key functions of balance assistive robots to enable the robot to accompany the user and provide balance support on time during gait training. To ensure the safety and transparency of HRI, the robot is required to have a fast and accurate response to human movements. Due to the unpredictable nature of human movements, especially for people with disability, it's challenging to use conventional control methods like PID to fulfill the aforementioned performance requirements. One way to predict human motion intention is to observe human movements and make use of human kinematics information such as body segment positions and velocities. Usually, human kinematics can be obtained through vision sensors (\cite{chalvatzaki2019learn}, \cite{zhu2022novel}), Laser ranger finder (LRFs) \cite{yan2021human}, Ultra-Wideband (UWB) \cite{xue2022uwb} and wearable sensors \cite{yuan2019uncertainty}. However, these sensors are not cost-effective or convenient to put on for daily use in the home and community scenarios. Moreover, the non-holonomic constraints of mobile robots lead to an essential difference in kinematics between human and mobile robots. In some human following applications, human kinematics are simplified as a unicycle model with constrained motions, such as the works in (\cite{scheggi2014cooperative} and \cite{chalvatzaki2018user}, which is not applicable to scenarios with free daily activities. Consequently, the control law is expected to absorb the kinematic differences between humans and robots to overcome the limitations of mobile robots.

In this paper, an assistive robot called mobile robotic balance assistant (MRBA) is developed to solve the above challenges and can be deployed in homes and communities for rehabilitation and assistance purposes. A passive and intrinsic transparent HRI which can enable 3-DoF natural pelvic movement is presented. In the control design of the user following task, we acknowledge the unconstrained nature of human kinematics and incorporate it into the modeling of the human-robot coupling system.  We adopt cost-effective rotary encoders on HRI to measure human positions. From the controller design perspective, we design an observer-based method to estimate human inputs to the system, and a nonlinear controller with disturbance rejection is proposed to achieve the graceful following of the user while absorbing the kinematic differences between the user and robot. Both the mechanical and controller designs are novel, and the effectiveness of the proposed algorithm is validated and compared with conventional PID controller. 

\begin{not1}
    When defining the variables, we follow the rule that matrices are denoted by capitalized bold letters, vectors are denoted by lowercase bold letters and scalars are denoted by lowercase non-bold letters. The identity matrix of dimension $n$ is denoted by $\textbf{I}_n$ and an $m{\times}n$ matrix with all elements of zero is denoted by $\textbf{0}_{m{\times}n}$.
\end{not1}

\section{System description and Problem Formulation}

To fill the gaps in balance robots for home and community use, we developed a compact and cost-effective robot called MRBA to provide assistance for post-rehabilitation patients to do ADLs. The core functions of MRBA include: 1. user following to enable the robot to accompany the user from behind constantly and 2. gait evaluation and balance assistance to monitor the user's gait patterns in real-time and provide necessary assistance or intervene a fall. In this section, the mechanical design of MRBA and user following control problem are described.
For more information about MRBA, a video demonstration can be found via the link (\url{https://youtu.be/RMAb-N2n82U}).

\subsection{Mechanical design}\label{sec:2.1}

\begin{figure}
\begin{center}
\includegraphics[width=8.4cm]{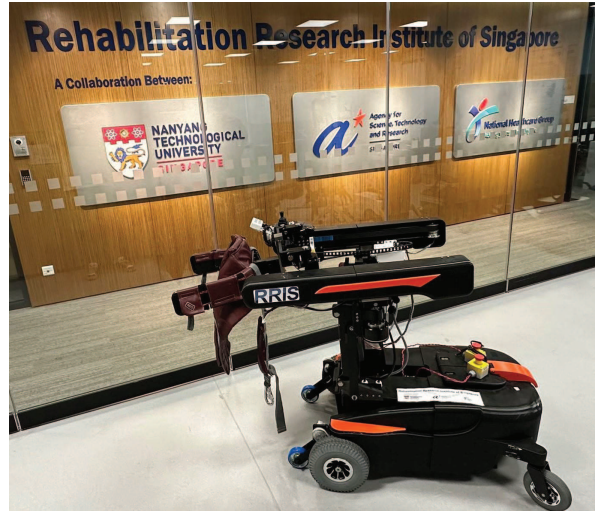}
\caption{Overview of the MRBA prototype} 
\label{fig:MRBA_concept}
\end{center}
\end{figure}

The MRBA consists of two main components: a 3-DoF robotic arm and a mobile wheelchair base as shown in Fig.~\ref{fig:MRBA_concept}. The robotic arm is designed to mimic the therapist's arm function to provide balance support at the pelvis. The powered wheelchair base enables the robot to navigate together with the user, and the robotic arm is connected to the wheelchair base through rotational joints.

The robotic arm holds a symmetrical parallel structure. In each half, there is a linear guide that contains a linear slider connected to the HRI. The HRI is attached to the user's waist through a belt to monitor the user's pelvic movement. Due to the intrinsic transparent property, the HRI enables compliant interaction and decouples the user dynamics from the mobile wheelchair base which reduces the impact of the robot on the user. In total, the robotic arm provides planar 3-DoF to realize the natural pelvic moment during the gait cycle. 

\begin{figure}
\begin{center}
\includegraphics[width=8.4cm]{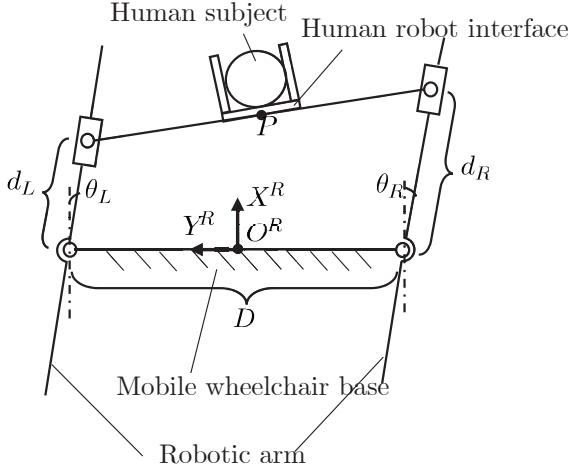}
\caption{Schematic diagram of the parallel robotic arm (top of view)} 
\label{fig:robotic_arm}
\end{center}
\end{figure}

Fig.~\ref{fig:robotic_arm} shows the geometric scheme of the robotic arm. $O^RX^RY^R$ is the coordinate fixed on the mobile base and the moving direction of the robot is along with $X^R$ axis. The fixed link indicates the mobile base. $d_L$ and $d_R$ are the variable distances between the left and right sliders and rotary joints respectively. $\theta_L$ and $\theta_R$ are the variable angles of the left and right arm with respect to the $X^R$ axis. Four cost-effective rotary encoders are adopted on the robotic arm to measure $d_L$, $d_R$, $\theta_L$ and $\theta_R$. $D$ is the fixed distance between the rotary joints of the two arms. $P$ is the center of HRI, which is equivalent to the human position in the robot coordinate. The orientation of HRI is equivalent to human orientation in the robot coordinate. Thus, the position and orientation of the human in the robot coordinate can be calculated with $d_L$, $d_R$, $\theta_L$ and $\theta_R$.

\subsection{System description}
To describe the system, we define the planar world and robot coordinate systems as $O^WX^WY^W$ and $O^RX^RY^R$, respectively. Let $^W\textbf{x}_r=[x_r, y_r, \theta_r]^{\top}$ and $^W\textbf{x}_h=[x_h, y_h, \theta_h]^{\top}$ be the state vectors of human and robot in the world coordinate, including the position and orientation information, $^R\textbf{x}_h=[x,y,\theta]^{\top}$ as the relative state of human in the robot coordinate which is measurable by the sensors on the robotic arm. The transformation of human states between world and robot coordinates can be represented as:
\begin{equation}\label{eq:1}
    ^R\textbf{x}_h= {_R^W\textbf{T}}({^W\textbf{x}_h}-{^W\textbf{x}_r}),
\end{equation}
where the transform matrix is:

\begin{equation}
    _R^W\textbf{T}=
\begin{bmatrix}
\cos{\theta_r} & \sin{\theta_r} & 0 \\
-\sin{\theta_r} & \cos{\theta_r} & 0 \\
0 & 0 & 1
\end{bmatrix}.
\end{equation}

The robot conforms to a unicycle model and the kinematics can be represented as:
\begin{equation}\label{eq:2}
    \Dot{x}_r=v_r\cos{\theta_r}, \, \Dot{y}_r=v_r\sin{\theta_r}, \, \Dot{\theta}_r=w_r,
\end{equation}
where $v_r$ and $w_r$ are the linear and angular velocities of the robot which are the control inputs in the system.

Different from the robot, human kinematics enable 3-DoF planar motion and can be represented as:
\begin{equation}\label{eq:3}
    \Dot{x}_h=v_h\cos{\delta}, \, \Dot{y}_h=v_h\sin{\delta},\, \Dot{\theta}_h=w_h,
\end{equation}
where $v_h$ and $w_h$ are the linear and angular velocities of the human. It is worth noticing that $\delta$ is the angle between $\dot{x}_{h}$ and $\dot{y}_{h}$, and $\delta$ is not same as $\theta_h$ necessarily.

By taking the first  derivative of \eqref{eq:1} over time and combining with \eqref{eq:2} and \eqref{eq:3}, the human-robot coupling system kinematics can be represented as:

\begin{equation}
    \begin{bmatrix}
    \Dot{x} \\ \Dot{y} \\ \Dot{\theta}
    \end{bmatrix}
    =
    \begin{bmatrix}
    yw_r-v_r+v_h\cos(\delta-\theta_r)\\
    -xw_r+v_h\sin(\delta-\theta_r)\\
    -w_r+w_h
    \end{bmatrix}.
\end{equation}

\begin{figure}
\begin{center}
\includegraphics[width=8.4cm]{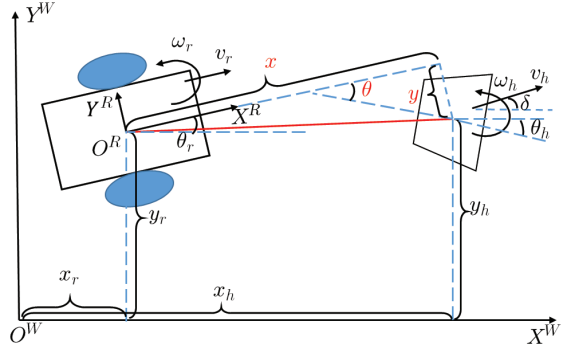}
\caption{Diagram of the human-robot coupling system (trapezoid: human, rectangular: robot).} 
\label{fig:system_diagram}
\end{center}
\end{figure}

Fig.~\ref{fig:system_diagram} shows the human-robot coupling system diagram. The robot should keep the fixed distance right behind the user like a therapist, which means the robot should keep the fixed distance $x_d$ along the $X^R$ axis and  $y_d$ along the $Y^R$ axis. To achieve the natural pelvis rotation of the user and allow the free relative orientation between the user and robot, the relative angle $\theta$ is not controlled and the kinematics of the system change to

\begin{equation}\label{eq:5}
    \begin{bmatrix}
    \Dot{x} \\ \Dot{y}
    \end{bmatrix}
    =
    \begin{bmatrix}
    yw_r-v_r+v_h\cos(\delta-\theta_r)\\
    -xw_r+v_h\sin(\delta-\theta_r)\\
    \end{bmatrix},
\end{equation}
where $v_h\cos(\delta-\theta)$ and $v_h\sin(\delta-\theta)$ are the components of human linear velocity along the $X^R$ axis and $Y^R$ axis, respectively. Since the human kinematics are unknown, these two parts can be considered as disturbances $d_x$ and $d_y$. Then, the system kinematics is transformed into:

\begin{equation}\label{eq:6}
    \begin{bmatrix}
    \Dot{x} \\ \Dot{y}
    \end{bmatrix}
    =
    \begin{bmatrix}
    yw_r-v_r+d_x\\
    -xw_r+d_y\\
    \end{bmatrix}
\end{equation}

Here, the system states $x$ and $y$ can be calculated based on the direct measurements of $d_L$, $d_R$, $\theta_L$ and $\theta_R$ through the sensors.

The problem of interest is designing the control inputs $v_{r}$ and $w_r$ such that the relative position of the human to the robot tends to the desired value $x_{d}$ and $y_{d}$ during the entire following process. 

\section{Controller Design}

\subsection{Feedback linearization}
Based on the nonlinear kinematics of the human-robot coupling system \eqref{eq:6}, the input-output feedback linearization \cite{henson1997nonlinear} is utilized to linearize the system kinematics.

\begin{def1}
Suppose \textbf{x} is an $n$-dimensional vector, \textit{h}(\textbf{x}) is a smooth scalar function and \textbf{\textit{f}}(\textbf{x}) is a smooth vector field. Then, $\nabla\textit{h}$ denotes the gradient of \textit{h} and $L_fh=\nabla\textit{h}\cdot \textbf{\textit{f}}$ denotes the Lie derivative of $h$ with respect to \textbf{\textit{f}}.
\end{def1}
Let $\textbf{x}=[x,y]^{\top}$, $\textbf{\textit{f}}(\textbf{x})=[f_1,f_2]^{\top}=[0,0]^{\top}$,  $\textbf{\textit{g}}(\textbf{x})=\begin{bmatrix}
g_1 & g_2
\end{bmatrix}=\begin{bmatrix}
-1 & y\\0 & -x
\end{bmatrix}$, $\textbf{u}=[v_r,w_r]^{\top}$, $\textbf{d}=[d_x,d_y]^{\top}$, $\textbf{y}=[y_1,y_2]^{\top}$, $\textbf{\textit{h}}(\textbf{x})=[h_1,h_2]^{\top}=[x,y]^{\top}$, system kinematics (\ref{eq:6}) can be transformed into a general form of affine state space model:
\begin{equation}
    \begin{cases}
    \Dot{\textbf{x}}=\textbf{\textit{f}}(\textbf{x})+\textbf{\textit{g}}(\textbf{x})\textbf{u}+\textbf{d}\\
    \textbf{y}=\textbf{\textit{h}}(\textbf{x})
    \end{cases}
\end{equation}
The relative degree vector of the system is $[
r_1,r_2]^{\top}
= [1,1]^{\top}$.
The decoupling matrices can be derived as:
\begin{equation}
    \textbf{M}(\textbf{x})=
    \begin{bmatrix}
    L_{g_1}L^{r_1-1}_fh_1 & L_{g_2}L^{r_1-1}_fh_1\\
    L_{g_1}L^{r_2-1}_fh_2 & L_{g_2}L^{r_2-1}_fh_2
    \end{bmatrix}
    =
    \begin{bmatrix}
    -1 & y \\ 0 & -x
    \end{bmatrix}
\end{equation}

\begin{equation}
    \textbf{b}(\textbf{x})=
    \begin{bmatrix}
    L^{r_1}_fh_1\\
    L^{r_2}_fh_2
    \end{bmatrix}
    =
    \begin{bmatrix}
    0\\0
    \end{bmatrix}
\end{equation}

Since $x$ maintains positive due to the mechanical constraints, the non-singularity of the decoupling matrix $\textbf{M}$ can be guaranteed.

To achieve the input-output decoupling, the control inputs $\textbf{u}$ are designed as:
\begin{equation}\label{eq:10}
    \textbf{u}
    =
    \textbf{M}^{-1}(\textbf{x})(\textbf{v}-\textbf{b}(\textbf{x}))
    =
    \begin{bmatrix}
    -v_x-\frac{y}{x}v_y\\-\frac{1}{x}v_y
    \end{bmatrix}
\end{equation}
where $\textbf{v}=[v_x,v_y]^{\top}$ is the transformed control inputs with respect to the linearized system. 

Then the linearized system kinematics can be derived from (\ref{eq:6}) as:
\begin{equation}\label{eq:11}
    \begin{bmatrix}
    \Dot{x} \\ \Dot{y}
    \end{bmatrix}
    =
    \begin{bmatrix}
    v_x+d_x\\
    v_y+d_y
    \end{bmatrix}
\end{equation}

\subsection{Observer-based human inputs estimation}
Considering human velocities are unknown inputs to the system that cannot be measured directly, we utilized observers to estimate human inputs.

Based on (\ref{eq:11}), this multi-input-multi-output (MIMO) system has been completely decoupled and can be separated into two single-input-single-output (SISO) systems. Here, two linear extended state observers (LESO) are implemented according to human inputs along $X^R$ and $Y^R$ axes separately \cite{han2009pid}.

Two LESOs are designed as:
\begin{gather}
    \begin{cases}
        \Dot{\hat{x}} = v_x+\hat{d}_x-\beta_1e_1 \\
        \Dot{\hat{d}}_x = -\beta_2e_1
    \end{cases}
\end{gather}

\begin{gather}
    \begin{cases}
        \Dot{\hat{y}} = v_y+\hat{d}_y-\beta_3e_2 \\
        \Dot{\hat{d}}_y = -\beta_4e_2
    \end{cases}
\end{gather}
where $\hat{x}$, $\hat{y}$, $\hat{d}_x$ and $\hat{d}_y$ are the estimations of $x$, $y$, $d_x$ and $d_y$ respectively, $e_1=\hat{x}-x$, $e_2=\hat{y}-y$. $\beta_1$, $\beta_2$, $\beta_3$ and $\beta_4$ are tuning observer gains. The convergence of the observers has been rigorously discussed by \cite{yoo2006convergence} and is omitted here.

\subsection{Nonlinear controller design}
To design the controller, we start by defining the tracking errors as $\textbf{e} = [e_x,e_y]^{\top} = [x - x_d,y-y_d]^{\top}$. Then, the dynamics of error states are derived from \eqref{eq:11} as:
\begin{equation}\label{eq:14}
\textbf{e}=\textbf{v}+\textbf{d}
\end{equation}

In order to eliminate the kinematics difference between the human and the robot, we utilize an optimization problem to minimize a linear quadratic cost function to balance the tracking errors and control inputs. The classic linear quadratic regulator (LQR) is an effective solution to the linear quadratic optimal problem but it cannot deal with the disturbances. According to \cite{dorato1994linear}, the LQR-based controller designed for systems with disturbances is expected to have two components: a feedback component to accord with optimal control in the absence of disturbances and a feedforward component to cancel out the effect of disturbances. Thus, the control inputs are split into two parts:
\begin{equation}\label{eq:16}
\textbf{v}=\textbf{v}_b+\textbf{v}_f
\end{equation}
where $\textbf{v}_b$ and $\textbf{v}_f$ are the feedback and feedforward components respectively. 

The dynamics of error state \eqref{eq:14} is equivalent to:
\begin{equation}\Dot{\textbf{e}}=\textbf{A}\textbf{e}+\textbf{B}_v(\textbf{v}_b+\textbf{v}_f)+\textbf{B}_d\textbf{d}, 
\end{equation}
with matrices $\textbf{A}=\textbf{0}_{2\times2}$, $\textbf{B}_v=\textbf{I}_{2}$ and $\textbf{B}_d=\textbf{I}_{2}$.

Then, the cost-to-go function associated with tracking errors and part of control inputs is designed as:
\begin{equation}
J=\frac{1}{2}\int_{0}^{\infty} (\textbf{e}^T\textbf{Qe}+\textbf{v}_b^T\textbf{Rv}_b)dt,
\end{equation}
where \textbf{Q} and \textbf{R} are positive definite matrices.

The solution to minimizing the cost function is given by:
\begin{equation}\label{eq:19}
    \textbf{v}_b=-\textbf{K}_e\textbf{e},
\end{equation}
where $\textbf{K}_e$ is given by:
\begin{equation}
    \textbf{K}_e=\textbf{R}^{-1}\textbf{B}^T_v\textbf{P}
\end{equation}

and $\textbf{P}$ is the solution to the continuous-time algebraic Riccati equation:
\begin{equation}
    \textbf{A}^T\textbf{P}+\textbf{PA}-\textbf{PB}_v\textbf{R}^{-1}\textbf{B}_v^T\textbf{P}+\textbf{Q}=0
\end{equation}

To compensate for the disturbances, the feedforward component is designed as:
\begin{equation}\label{eq:22}
    \textbf{v}_f=-\textbf{B}_v^+\textbf{B}_d\hat{\textbf{d}},
\end{equation}
where $\textbf{B}_v^+$ is the Moore–Penrose pseudoinverse of $\textbf{B}_v$ and $\hat{\textbf{d}}=[\hat{d}_x,\hat{d}_y]^{\top}$ is the estimation of $\textbf{d}$. 

Consequently, the overall control inputs are designed by substituting (\ref{eq:19}) and (\ref{eq:22}) into (\ref{eq:16}):

\begin{equation}
    \textbf{v}=-\textbf{K}_e\textbf{e}-\textbf{B}_v^+\textbf{B}_d\hat{\textbf{d}}
\end{equation}

The actual inputs to the system can be achieved according to (\ref{eq:10}):

\begin{equation}\label{eq:24}
    \textbf{u}=\begin{bmatrix}
    v_r\\w_r
    \end{bmatrix}
    =
    \textbf{M}^{-1}(\textbf{v}-\textbf{b})
    =
    \textbf{M}^{-1}(-\textbf{K}_e\textbf{e}-\textbf{B}_v^+\textbf{B}_d\hat{\textbf{d}}-\textbf{b})
\end{equation}

Since the linearized system has been decoupled into two SISO systems, $\textbf{Q}$ and $\textbf{R}$ can be chosen as diagonal matrices without losing generality. As a result, $\textbf{K}_e$ to be determined is diagonal as well: 
\begin{equation}
    \textbf{K}_e = \begin{bmatrix}
k_x & 0\\0 & k_y
\end{bmatrix}
\end{equation}

The nonlinear control inputs are given in the explicit form as
\begin{equation}
    \textbf{u}=\begin{bmatrix}
    v_r\\w_r
    \end{bmatrix}
    =
    \begin{bmatrix}
    \frac{y}{x}(k_{y}e_y+\hat{d}_y)+k_{x}e_x+\hat{d}_x \\ \frac{1}{x}(k_{y}e_y+\hat{d}_y)
    \end{bmatrix}
\end{equation}

\section{Results}

In this section, we implemented the proposed controller on MRBA to perform the user-following task. The experiment with MRBA has been done to collect voltage and velocity data for system identification. We took the real system dynamics into account when validating our controller. In addition, realistic human trajectory data which were collected by the \cite{qualisys2008qualisys} motion capture system (Qualisys Miqus M3 (2MP)) were used for performance evaluation.
\
\subsection{Model validation}

The mobile base of MRBA is powered by a battery with a maximum voltage of 24 V. The two wheels are driven by 2 brushless DC motors and on each wheel, there is an absolute rotary encoder with a resolution of 4096 pulses per revolution.

\begin{figure}
\begin{center}
\includegraphics[width=8.4cm]{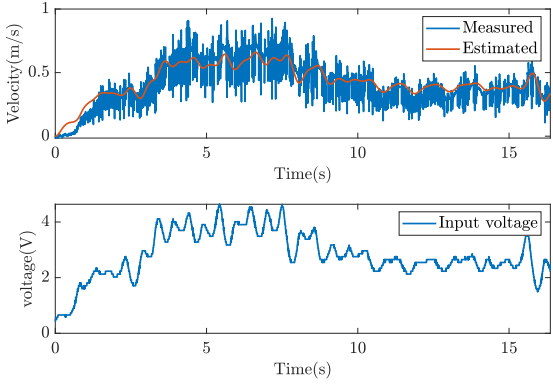}
\caption{Model validation results. Upper: measured and estimated velocities of left wheel; Lower: input voltage.} 
\label{fig:systemid}
\end{center}
\end{figure}

To validate the proposed control algorithm in high-fidelity simulation, we conducted the experiment of MRBA to identify and validate the dynamics of the system. The system parameters were estimated by the nonlinear least-square method with collected input voltage and wheel velocity data. Model validation was conducted by comparing the measured wheel velocities and estimated wheel velocities from the identified dynamic model. The model validation result is shown in Fig.~\ref{fig:systemid} and it can be seen that the estimated velocity follows the trend of measured velocity given the same input voltage.

\subsection{Performance evaluation}

Then, we collect realistic human pelvic motion trajectories during walking with the Qualisys motion capture system. The target subjects have the ability of overground walking without active assistance but withstand higher risks of fall than normal people. Four markers were placed symmetrically surrounding the subject's waist and the geometric centers of these four markers were chosen as the reference point that the robot is expected to follow. To cover different gait patterns in real applications, the subject was required to walk in different modes like moving in place, continuously changing directions, accelerating and decelerating during data collection.

The sampling frequency $f_s$ is fixed at \SI{200}{\hertz}. For human inputs estimation, we chose $\beta_1=\beta_3=1$ and $\beta_2=\beta_4=\frac{1}{3}f_s=\frac{200}{3}$ for the LESO implementation based on the suggestions given by \cite{han2009pid}. For parameter settings in the nonlinear controller, we chose $\textbf{Q}=\begin{bmatrix}
200 & 0\\0&200
\end{bmatrix}$ and $\textbf{R}=\textbf{I}_2$ and the corresponding controller parameters are $k_x=14.1421$ and $k_y=14.1421$. 

\begin{figure}
\begin{center}
\includegraphics[width=8.4cm]{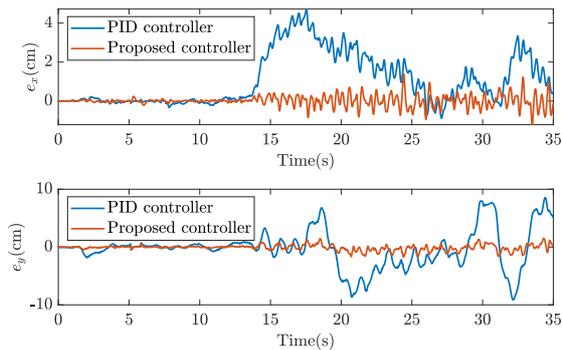}
\caption{Tracking errors comparison between PID and proposed controllers. Upper: tracking errors in $x^R$; Lower: tracking errors in $y^R$.} 
\label{fig:trajectory}
\end{center}
\end{figure}

Fig.~\ref{fig:trajectory} shows the trajectories of the human and robot with the proposed controller in the world coordinates respectively. The robot can follow the human trajectory quite well in different walking modes.  
\begin{figure}
\begin{center}
\includegraphics[width=8.4cm]{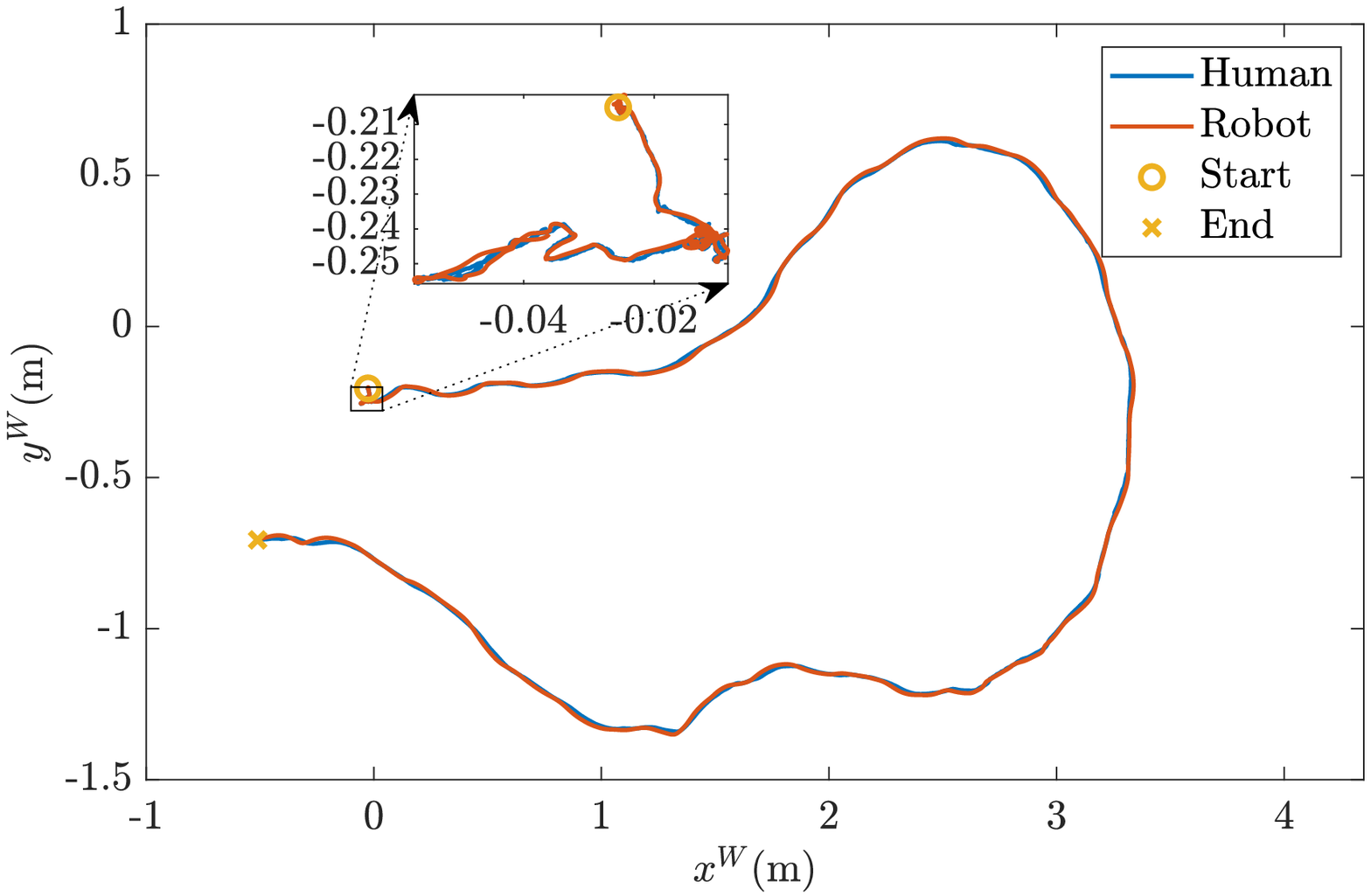}
\caption{Human and robot trajectories in world coordinates. The zoom-in zone depicts the trajectories of human and robot when moving in place before beginning to walk.} 
\label{fig:tracking_error}
\end{center}
\end{figure}

For comparison, the conventional PID controller was implemented as a benchmark for performance evaluation. Tracking errors of PID and proposed controllers along $X^R$ and $Y^R$ axes are shown in Fig.~\ref{fig:tracking_error}. In the beginning, the subject kept moving in place and both PID and the proposed controller had good tracking performance. However, when the subject began to walk, the tracking errors of PID controller increased intensively especially when the subject tried to change directions frequently or accelerate and decelerate while the tracking errors of the proposed controller maintained small all the time. In total, the average and maximum absolute values of tracking errors based on PID along $X^R$ and $Y^R$ axes after beginning to walk are given in Tab.~\ref{tb:ErrorTab}. The average absolute values of errors reduce 84.36\% and 85.42\% in $X^R$ and $Y^R$ respectively and the maximum absolute values of errors reduce 70.34\% and 81.11\% in $X^R$ and $Y^R$ respectively of proposed controller compared to PID controller. The results have shown significant improvements in the proposed controller in tracking performance than the conventional PID controller. 

In clinical scenarios, the user can have sudden changes in their velocities to achieve ADLs. The failure to maintain the small tracking errors in terms of these sudden changes in velocities by the PID controller will increase the interaction force between the user and HRI, which will make the user feel uncomfortable and restrict the user's actions to some extent. However, the proposed controller can deal with sudden changes in user velocities leverage of real-time estimation and compensation of human inputs to the system, which will increase the transparency of HRI.

\begin{table}[hb]
\begin{center}
\caption{Statistical analysis of tracking errors}\label{tb:ErrorTab}
\begin{tabular}{ccccc}
\hline
& \multicolumn{2}{c}{PID controller} & \multicolumn{2}{c}{Proposed controller}\\
\cline{2-5}
& $\lvert e_x \rvert$ (cm) & $\lvert e_y \rvert$ (cm) & $\lvert e_x \rvert$ (cm) & $\lvert e_y \rvert$ (cm) \\
\hline
Avg. & 1.79 & 3.43 & 0.28 & 0.50 \\
Max. & 4.72 & 9.16 & 1.39 & 1.73 \\
\hline
\end{tabular}
\end{center}
\end{table}

\section{Conclusion}
This paper introduces the MRBA: a mobile robotic balance assistant for gait training and balance support. To reduce the inertia transmitted from HRI, a passive intrinsic transparent HRI was designed. Since MRBA can accompany the user and serve as a therapist, we proposed a graceful user-following method. An explicit mathematical model of the human-robot coupling system was constructed. Human inputs which are considered unknown disturbances were estimated by the observer-based method. Finally, an LQR-based controller with disturbance rejection was proposed. The effectiveness of the proposed method has been shown in high-fidelity performance evaluation with real system dynamics models and realistic human trajectories.

\begin{ack}
The authors would like to thank Jiaye Chen and Youlong Wang from the Rehabilitation Research Institute of Singapore for their support in the experiment setup.

The authors also would like to thank Christopher Wee Keong Kuah and Huiting Zhuo from Tan Tock Seng Hospital for their advice from a clinical perspective and Dr Ye Wang from the University of Melbourne for his active and constructive discussions.
\end{ack}

\bibliography{ifacconf}             
                                                  
\end{document}